%% file: acl21-sustainability-frame.tex
\documentclass[11pt,a4paper]{article}
\usepackage{authblk}
\usepackage{breakurl}
\usepackage{hyperref}
\hypersetup{breaklinks}
\usepackage[hyperref]{acl2021}
\usepackage{times}
\usepackage{latexsym}

\usepackage{graphicx}

\usepackage{enumitem}
\usepackage{booktabs}
\usepackage{multirow}

\usepackage{siunitx}
\usepackage{placeins}
\usepackage{stfloats}
\usepackage{makecell}

\usepackage{microtype}

\aclfinalcopy 
\def\aclpaperid{2826} 

\usepackage{adjustbox}
\usepackage{array}
\usepackage{setspace}

\newcolumntype{R}[2]{%
	>{\adjustbox{angle=#1,lap=\width-(#2)}\bgroup}%
	l%
	<{\egroup}%
}

\newcommand{\tablespacing}{\\[2pt]\\[-3em]}

\newcommand{\Ni}{(1)~}
\newcommand{\Nii}{(2)~}
\newcommand{\Niii}{(3)~}

\usepackage{subcaption}

\usepackage{graphicx}
\graphicspath{{acl21-sustainability-figures/}}
\DeclareGraphicsExtensions{.pdf,.ai,.jpg,.png}
\setkeys{Gin}{pagebox=artbox}

\begin{document}
  
\input{acl21-sustainability-pre}
\input{acl21-sustainability-part1}
\input{acl21-sustainability-part2}
\input{acl21-sustainability-part3}
\input{acl21-sustainability-part4}
\input{acl21-sustainability-part5}
\input{acl21-sustainability-part6}
\input{acl21-sustainability-sum}

\section*{Acknowledgments}

We thank the anonymous reviewers for their valuable and constructive feedback. We also thank the LMBV for many interesting and fruitful discussions.
This research was partially funded by the Development Bank of Saxony (SAB) under project numbers 100335729 and 100400221.

\input{acl21-sustainability-ethics}

\bibliography{acl21-sustainability-lit}
\bibliographystyle{acl_natbib}

\appendix
\input{acl21-sustainability-supplementary}

\end{document}

%% file: acl21-sustainability-pre.tex
\setlength\titlebox{8.1cm}

\title{Supporting Land Reuse of Former Open Pit Mining Sites\\using Text Classification and Active Learning}

\author[1,5]{\bf Christopher Schr\"oder}
\author[1,4,5]{\bf Kim B\"urgl}
\author[2,5]{\bf Yves Annanias}
\author[1,5]{\bf Andreas Niekler}
\author[1,4,5]{\\\bf Lydia M\"uller}
\author[2,5]{\bf Daniel Wiegreffe}
\author[3,5]{\bf Christian Bender}
\author[3,5]{\bf Christoph Mengs}
\author[2,4,5]{\\\bf Gerik Scheuermann}
\author[1,4,5]{\bf Gerhard Heyer}
\affil[1]{Natural Language Processing Group}
\affil[2]{Image and Signal Processing Group}
\affil[3]{Institute of Public Finance and Public Management}
\affil[4]{Institute for Applied Informatics (InfAI), Leipzig, Germany}
\affil[5]{Leipzig University, Germany\authorcr\texttt{\{schroeder,buergl,annanias,aniekler\}@informatik.uni-leipzig.de}\authorcr\texttt{\{lydia,daniel,scheuermann,heyer\}@informatik.uni-leipzig.de}\authorcr\texttt{\{bender,mengs\}@wifa.uni-leipzig.de}}
\date{}

\maketitle
\begin{abstract}
Open pit mines left many regions worldwide inhospitable or uninhabitable. Many sites are left behind in a hazardous or contaminated state, show remnants of waste, or have other restrictions imposed upon them, e.g., for the protection of human or nature. Such information has to be permanently managed in order to reuse those areas in the future. In this work we present and evaluate an automated workflow for supporting the post-mining management of former lignite open pit mines in the eastern part of Germany, where prior to any planned land reuse, aforementioned information has to be acquired to ensure the safety and validity of such an endeavor. Usually, this information is found in expert reports, either in the form of paper documents, or in the best case as digitized unstructured text---all of them in German language. However, due to the size and complexity of these documents, any inquiry is tedious and time-consuming, thereby slowing down or even obstructing the reuse of related areas. Since no training data is available, we employ active learning in order to perform multi-label sentence classification for two categories of restrictions and seven categories of topics. The final system integrates optical character recognition (OCR), active-learning-based text classification, and geographic information system visualization in order to effectively extract, query, and visualize this information for any area of interest. Active learning and text classification results are twofold: Whereas the restriction categories were reasonably accurate {($>$0.85~F1)}, the seven topic-oriented categories seemed to be complex even for human annotators and achieved mediocre evaluation scores ($<$0.70~F1).
\end{abstract}

%% file: acl21-sustainability-part1.tex
\section{Introduction}

In many parts of the world, raw materials were mined in open pit mines during the last century, leaving many of these regions inhospitable or uninhabitable. To put these regions back into use, entire stretches of land must be {\em renaturalized}, which means that land must be ecologically restored with the aim to ultimately increase biodiversity, or {\em recultivated}, which means its productivity must be restored, e.g., reused for agriculture, recreational areas, industrial parks, solar and wind farms, or as building land \cite{luc:2015}. In the following, we subsume both renaturalization and recultivation under {\em land reuse}. For land reuse, it is essential that all relevant information about the sites is retained, which used to be recorded in the form of textual reports. Such reports include information such as, among others, hazards, soil composition, or environmental factors. Therefore, having access to all these reports, it can be determined if a site can be reused immediately, only under certain conditions, or not  at all in the foreseeable future. 

For reaching a sustainable future, the \citet{un:2015} has defined objectives, called sustainable development goals (SDGs). Land reuse is a shared common denominator among several of those goals such as ``Zero hunger'', ``Clean water and sanitation'', ``Sustainable cities and communities'', ``Climate action'', ``Life below water'', and ``Life on land''. Moreover, it provides co-benefit to all SDGs as shown by \citet{herrick2019land} and directly supports ``Life on Land''. By implication, anything that obstructs land reuse also impedes the fulfillment of several SDGs.

\noindent This work deals with the real-world use case of post-mining management \cite{kretschmann:2020} of former lignite open pit mines in the eastern part of Germany. Here, a large number of such documents exist, and moreover, there is metadata maintained, which maps each document to its related area. Apart from that, before any land can be reused in these areas, it is legally required that local authorities must be consulted before proceeding any further. This process includes seeing through numerous legacy documents, which is laborious, time-consuming and delays a subsequent reuse of such areas. We address this issue by demonstrating and evaluating a workflow consisting of optical character recognition (OCR), text classification and active learning, whose results are then visualized by a Geographic Information System (GIS). By automating information extraction and making extracted results available through a GIS, we increase efficiency by which information about a specific location of interest can be queried. This can accelerate the reuse of land by supporting the efficiency of employees managing these areas, and thereby contributes towards the fulfillment of several SDGs.

This necessary review of a multitude of documents, which is obligatory prior to any land reuse, is aggravated even more by Germany's federal structure \cite{gov:2016} due to which land management is a task of the municipalities. The federal government, as well as the states are responsible for the SDGs' implementation, which is then passed on to the municipalities, which therefore are effectively responsible for supporting SDGs. Municipalities, however, do not have a standardized software infrastructure \cite{Zern-Breuer:20}, which results in a heterogeneous data management landscape and thereby makes the implementation of SDGs challenging, especially for small municipalities. Information about former lignite open pit mines is stored in independent GISes, related unstructured documents are stored in dedicated storage systems (either in form of piles of paper, scanned documents, or even as digitized text), and the connections between documents and geographic coordinates are stored in yet other databases. In order to obtain information about an area of interest, all information must be contextualized, compiled, and manually evaluated. 

Although the presented approach is tailored towards the post-mining management in Eastern Germany, this is relevant to many other countries in the world, which are also concerned with stopping lignite and coal mining to reduce $\text{CO}_2$ emissions. To give a few examples, Belgium performed coal phase-out in 2016, Sweden and Austria in 2020; Canada will follow in 2030, and Germany in 2038. All of these countries will need to post-manage former mining sites in order to reuse the affected areas. Apart from lignite and coal mining, this is also true for other mining sites. Once resources are exhausted or are no longer needed, land has to be renaturalized or recultivated or will stay deserted for unknown time. 

%% file: acl21-sustainability-part2.tex
\section{Foundations and Related Work}

Sustainability issues have long been politically ignored, but became much more relevant in recent years. As a result, this societal challenge has recently started to get traction in computer science \cite{gomes:2019} and  natural language processing \cite{conforti:2020}, where only few previous works study methods to support SDGs: \newcite{conforti:2020} classify user-perceived values on unstructured interview text in order to gather structured data about the people's subjective values. This is performed in developing countries to increase the success of sustainability projects, each targeted at one or more SDGs, by aligning them to the encountered values, so that the projects will be more likely to be continued by the community after their initial implementation. Similar to us, they also performs sentence classification to support SDGs, however, besides using data from a completely different domain, we perform multi-label classification, use more recent transformer-based models, and integrate additional geospatial information. \citet{pincet} support the automatic classification of SDGs in reporting documents in the form of an official API for the OECD (Organisation for Economic Co-operation and Development), which is responsible for implementing SDGs and monitoring the progress thereof. This clearly shows the problem of an increasing number of documents relevant for implementing SDGs, and also the need for tools to support such processes.

There are a variety of OCR engines available, with  Tesseract~\cite{Smith1987} being a good starting point. Tesseract offers a number of pre-processing mechanisms for document images, however, it does not implement the full range of  state-of-the-art OCR. Image pre-processing as proposed and implemented by the OCR-D project~\cite{Binmakhashen,neudecker}, is beneficial to additionally extend the tool with the latest developments in OCR.

\indent In recent years, text classification, like many other fields in natural language processing, has experienced a paradigm shift towards transformer-based models \cite{vaswani:2017,devlin:2019}, which raised the state-of-the-art results on many tasks. Besides the impressive performance gains, the main advantage of using a pre-trained model is that its performance can be translated to low-data scenarios, which were previously challenging due to deep models overfitting on small data. Transformers, however, have been shown to work well on small data \cite{ein-dor:2020,yuan:2020}, and consequently open up new possibilities on previously challenging tasks.

\indent Geographic information systems are a common technological choice to visualize spatial data on cartographic maps and have been shown to be invaluable for supporting SDGs \cite{Avtar:20}. Using a GIS, one can combine a database storing textual information with the spatial data to support experts in the decision-making process or to enable the exploration of data.
To support renaturalization of a river valley, \citet{matysik2012renaturization} used a GIS to analyze hydrological aspects in the area and develop a plan for renaturalization. Similarly to our work, they also combined several layers of features in the GIS. A recent toolbox of the commercial ARCGIS software called LocateXT \cite{locatetx} can connect a larger number of unstructured datasets into a running GIS, however, although it can automatically link information and coordinates from the data, it does not support the extraction and processing of unstructured information and other attributes providing further information.

%% file: acl21-sustainability-part3.tex
\section{Data}
\label{sec:data}

\begin{figure*}[tb]
    \centering
    \includegraphics[width=0.9\textwidth]{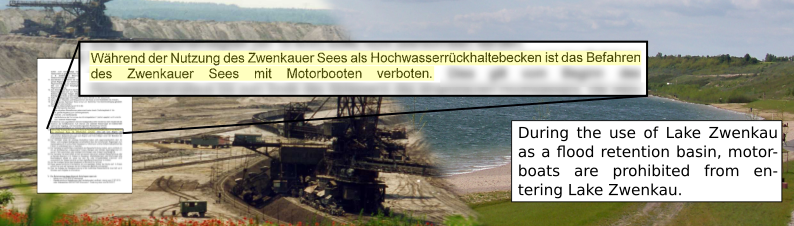}
    \caption{Example of a typical documentation. Part of the textual reports are passages about restrictions or prohibitions in the described area. (The background image consists of two photos, one by \href{https://flickr.com/people/sludgeulper/}{sludgeulper} (\href{https://commons.wikimedia.org/wiki/File:Schaufelradbagger,_Braunkohle_Tagebau_-_Open_Cast_Lignite_mining,_Espenhain_near_Leipzig,_June_1993_(4160611427).jpg}{left background}, \href{https://creativecommons.org/licenses/by-sa/2.0/deed.en}{CC BY-SA 2.0}), and the other by \href{https://commons.wikimedia.org/wiki/User:Joeb07}{Johannes Kazah} (\href{https://commons.wikimedia.org/wiki/File:Markkleeberger_See_Strand.jpg}{right background}, \href{https://creativecommons.org/licenses/by-sa/2.0/deed.en}{CC BY-SA 2.0}). The resulting image changes the originals only by adding overlays (to the front) and is also licensed under the \href{https://creativecommons.org/licenses/by-sa/2.0/deed.en}{CC BY-SA 2.0} license.)}
    \label{fig:zwenkau}
\end{figure*}

We use data from the Lausitzer und Mitteldeutsche Bergbauverwaltungsgesellschaft mbH (LMBV)\footnote{\url{https://de.wikipedia.org/wiki/LMBV}}, who are responsible for the management and reuse of abandoned mining sites in the eastern part of Germany. For this purpose, they archive and manage all documents related to sites in this area, and issue new documents if required. Moreover, they are obligated to provide reliable information about the managed lands for the public on {\em request}. Such requests require, among others, to inform about any restrictions for the specific area, which can be found in the associated documents. An illustrated example is shown in Figure \ref{fig:zwenkau}. 

For research, the LMBV provided us with 31,605 of such documents (16,883 for the region Lausitz\footnote{\url{https://en.wikipedia.org/wiki/Lusatia}}, 14,722 for the region Mitteldeutschland\footnote{\url{https://en.wikipedia.org/wiki/Central_Germany_(cultural_area)}}). The oldest documents date back to the 1960s, but scans were only produced within the last 20 years. The documents encompass several different types, for example, reports, drilling logs, expert opinions, statements, plans, maps, and correspondences. The quality of the scans varies from excellent to fair quality. Moreover, some documents are stored in other digital formats (.doc, .docx, .odf) and others are stored as scanned images. The documents have different origins: They are authored by the companies mining the open pit mine, by the LMBV managing the closed open pit mines, by companies responsible for certain subtasks such as building infrastructure, or by other experts. They include documents from the time when  open pit mines were actively mined but also documents created after the mines were closed. Besides the documents, our dataset contains over 30,000 geographic features. These features are described as points, lines, polygons and multipolygons, and can be visualized in a GIS. In addition, these data are provided with additional non-spatial information, such as the geographical affiliation of the documents mentioned. 
\subsection{Labels}

In this work, the goal is to find restrictions and topics, which are described in Table~\ref{tab:labels} and which will be used as labels during text classification. Restrictions are formulated in many different ways, e.g., a specific action may be forbidden, an action may require specific preceding steps to be allowed, or the action may be explicitly allowed under certain circumstances. Moreover, a restriction may refer to certain topics, e.g., restricting a construction method depending on the weather. Regarding topic labels, due to the different types of documents, they vary largely. For example, geotechnical issues can be frequently found in experts' opinions from geotechnical experts but may also appear in reports, statements or correspondence. Thus, topics are not limited to certain type of document and within one document or even one sentence more than one topic may appear. Likewise, restrictions can be found in most types of documents and describe known issues with the associated area. Those labels are deductively defined and reflect the requirements of the most frequent requests to the LMBV. Since this label system is specifically defined for this novel approach, no training or pre-labeled data could be provided by the LMBV, but for each label example sentences and common keywords were defined.\\

\subsection{OCR}\label{sec:ocr}

For documents which are not digitized yet, the text is extracted using Tesseract\footnote{\url{https://github.com/tesseract-ocr/tesseract}} and best practices regarding German language from the OCR-D community~\cite{Smith1987,neudecker,Binmakhashen}. The major challenges here were: \Ni The vast number of documents make it infeasible to optimize OCR parameters for each document, therefore OCR has to be optimized with regard to the whole collection. \Nii There is no manually transcribed evaluation data. \Niii The documents are written by humans without any review process making erroneous words or grammar very likely. For these reasons, and because of the many varying document types, investigating OCR quality is impractical and therefore outside the scope of this work. However, we use the built-in Tesseract evaluation procedure to judge the overall quality of the process and apply further filtering to cope with difficult documents and insufficient OCR quality.

OCR pre-processing steps for the images included orientation analysis and rotation, resizing of the image (400dpi), denoising, lighting intensity correction, binarization, and deskewing. Lighting intensity correction only improved the result in some cases but worsened the result in others. It is therefore only used if it improves the result based on the confidence score from Tesseract as explained below. Denoising converts the images into grayscales, applies a dilution filter, an erosion filter, and finally, a median blur filter.
 
The quality of the results is measured by evaluating the confidence score as produced by Tesseract which provides a word level confidence score reflecting the OCR quality. We aggregated the word level confidence score to a page level confidence score by averaging over all recognized words, resulting in a score between 0\ and 100\%, and assigning pages without recognized text a score of 0\%. Again, the document quality and layout is very heterogeneous and annotating a test set for the OCR process would lack completeness. We identified 45,141 (Mitteldeutschland) and 35,256 (Lausitz) pages in the document dataset. We accept all pages for our experimental dataset which are evaluated with a confidence score of more than 75\%. Without pre-processing only 45\% of the pages are detected with a confidence score of at least 75\%. The correct pre-processing improved the OCR result to 97\% of pages exceeding the defined threshold for the region Lausitz (from 44\% to 93\% for region Mitteldeutschland, respectively). In 93\% (Lausitz) and 83\% (Mitteldeutschland) of the pages with a confidence below 75\% the original documents do not contain any recognizable text. Hence, the majority of unrecognized or insufficiently recognized documents does not contain proper amounts of text.

\input{table-labels}

\subsection{Datasets and Splits}

In order to obtain both a point of reference for evaluation and an initial set of labeled data for the initial active learning model, we manually labeled a subset of 2000 sentences. For each label, we defined a set of keywords (see Appendix Table \ref{tab:keywords}), which are used to find sentences in the unlabeled data, that likely belong to that label. This is necessary because of the high ratio of unlabeled to labeled sentences in most documents, i.e., a majority of the sentences in the complete dataset will not have any label assigned. Keywords were used to locate restrictions and prohibition candidates. From this candidate pool, we select candidates for the topic categories utilizing further keyword matching. In doing so, we take a maximum of 150 examples per topic category. If no more than 300 candidates can be found for a topic category, we only include half of them in the candidate list to leave examples of such rare categories for active learning in our unlabeled dataset. Since we want to demonstrate the capabilities of active learning this is a necessary decision. Additionally, we added more than 700 randomly drawn sentences, resulting in a dataset of 2000 sentences in total. This dataset was annotated by three different (non-expert) annotators with the help of a guideline describing each label. We measured the inter-annotator agreement with Krippendorff's $\alpha$ \cite{krippendorff2011agreement} which resulted in values between 0.91 and 0.7, with \emph{Restricted Area} as the most agreed label and \emph{Construction} the least agreed label between annotators. This confirms our observation that labeling in this domain is challenging and needs domain expertise.

\indent We combine the annotations of all annotators by majority voting in order to obtain more stable judgments of our non-expert annotators \cite{nowak2010reliable}. \emph{``Requirement''} is the most frequent label, while \emph{``Weather''} is the least frequent. The true label distribution is unknown, but at least some of the labels seem to occur very rarely. We split the annotated dataset into training (500 samples), validation (500 samples), and test (1000 samples) set using iterative stratification \cite{sechidis2011stratification} to preserve the label distribution in all three sets (see Appendix, Table \ref{tab:label_distribution}).

\subsection{Geospatial Connection}  

The linkage of the non-spatial data (i.e., the documents and predicted restriction and topic labels) and the spatial data (e.g., coordinates for certain areas) can be represented as a graph. For this, the documents and the associated areas are expressed as nodes. Then, edges are used to link these document nodes to their corresponding area node. The graph serves as an efficient data structure, which is necessary to make the data available in a GIS and to enable the answering of requests by linking and collecting the required information.

\indent The predicted labels can be integrated into the data model with the following procedure:
For each topic (see Table~\ref{tab:labels}), an additional node is created carrying the label description as a node property. Restrictions that have not been classified more precisely by a topic are grouped together under a generic topic node. Then, edges are created for each restriction, pointing from the associated topic node to the document node from which the restriction originated. Additional information about the restrictions is available as attributes of the respective edges. This includes the sentence from which the restriction is derived and the confidence value from the text classification algorithm. These attributes are attached  to the edge instead of the document node itself, since a document can lead to several restrictions either related to the same topic or a different topic (e.g., ``{\em large installations may not be built}'' [construction-related] and ``{\em may not enter shore areas during heavy rain}'' [weather, restricted area]). 
Thus, a document node may be connected to (one or more) topic nodes via several edges that contain more detailed information about restriction and the corresponding sentence.

\indent Many queries can be realized with this data structure. For example, it is possible to efficiently query which restrictions exist in the same topic, in the same document, or in the same geographic area, since only the corresponding nodes need to be followed in the data model. This enables, in particular, an exploratory search that incorporates information from existing projects that may be relevant for a given request (see the use case in Section~\ref{sec:vis_usecase}).

%% file: table-labels.tex
{
\renewcommand{\arraystretch}{1.4}
\begin{table*}[t]
    \centering
    \small
    \begin{tabular}{p{0.2cm} l p{6cm} p{6cm}}
        \toprule
        & \textsc{\bfseries{}Label} & \textsc{\bfseries{}Description} & \textsc{\bfseries{}Example}\\
        \hline
        \tablespacing
        \multirow{2}{*}[-0.15em]{\rotatebox{90}{\small\textbf{Restrictions}}} & \textbf{Prohibition} & Statements which actively prohibit or restrict actions in general or conditionally. & \emph{Machines heavier than 30t are forbidden, landslide hazard}.\\
        & \textbf{Requirement} & Requirements limit usages and/or are directives what is to be done. & \emph{The area must be secured with 'no trespassing'-signs}.\\
        \tablespacing
        \hline
        \tablespacing
        \multirow{7}{*}[-3.2em]{\rotatebox{90}{\small\textbf{Topics}}} & \textbf{Weather} & Weather-related phenomena, consequences, and protection measures. & \emph{Shore areas must be avoided during heavy rain}.\\
        & \textbf{Construction} & Statements about construction plans, construction sites, or  construction procedures. &  \emph{Only one-storey buildings should be placed around the marina.}\\        
        & \textbf{Geotechnics} & Information related to the ground, e.g., about soil, stability, or  slopes. & \emph{Slopes must be protected against the effects of the weather.}\\      
        & \textbf{Restricted area} & Indicates a limited accessibility, mostly due to hazards, soil stability, or safety precautions. & \emph{Always keep a distance of at least 50m to the shore.}\\  
        & \textbf{Planting} & Plans, reports, or specific details about the type of plant and location of plantings. & \emph{Native species of bushes must be planted on the slope, to stabilize it against rupture.}\\   
        & \textbf{Environment} & For renaturalization, it is often strictly regulated where to plant, what, types etc. & \emph{Forest operations are forbidden during breeding season.}\\   
        & \textbf{Disposal} & Instructions concerning storage and disposal of (building) materials. & \emph{Contaminated soil must be cleaned and provably be disposed of.}\\   
        \bottomrule
    \end{tabular}
    \caption{Description of restrictions and topics, illustrated by examples (translated from German into English).}
    \label{tab:labels}
\end{table*}
}

%% file: acl21-sustainability-part4.tex
\section{Approach}
 
The goal of our approach is to detect restriction and topic labels at the sentence level. Subsequently, we can map predicted labels to geospatial data, which is already available in a structured format. This means, with a process chain of OCR, text classification, and GIS, we can effectively detect the presence or absence of labels at geographic coordinates of interest. In the end, this can be directly used to manage land reuse efforts, thereby supporting aforementioned SDGs. As existing OCR solutions are tried and tested, and the geospatial link is already given, the main challenge of this method is text classification,  namely: \Ni There is no predefined industrial standard for the labels which are not formally defined but given by some exemplary formulations and keywords provided by the LMBV. Consequently, the definitions are incomplete and new formulation not using the keywords are expected. \Nii The documents exhibit a domain-specific, often convoluted, vocabulary.

\subsection{Text Pre-processing}\label{sec:data_processing}

Since the following text classification depends on the quality of the raw text obtained through the OCR step, which we observed to be rather noisy owing to the structure of some documents, we applied a series of pre-processing steps: We detect word wraps and remove the hyphen, convert line breaks into white space, and finally trim repeated sequences of white space. Subsequently, sentence segmentation was performed using syntok\footnote{\url{https://github.com/fnl/syntok}}. In order to filter out sentences which are obviously erroneous, e.g., sentences containing only gibberish words, we filtered all sentences which violate the properties of a valid sentence~\cite{goldhahn:2012}. This was achieved by a set of regular expressions and filter rules, which detect improper sentences, e.g., sentences which contain too many special characters, start with a lowercase letter, or are missing a terminal punctuation character.

\subsection{Text Classification and Active Learning}\label{sec:SentenceClassification}

Using the extracted sentences described in Section \ref{sec:data_processing} as input, our goal is to classify restriction and topic labels. In contrast to standard text classification datasets, the LMBV data, like most real-world data, provides no labels. Manually labeling documents, however, is time-consuming and therefore costly, especially when some labels are very rare. For this reason, we use active learning \cite{lewis:1994}, which works as follows: In an iterative process the active learner presents unlabeled data to a user, which the user has to label. The purpose of this is to reduce the total labeling effort, by identifying samples that add the most value to the current model. The key for this is the query strategy, which selects examples to be labeled by the user.  After labeling the presented samples, a new model is trained, and the loop is repeated, either for a specific number of rounds, or until a stopping criterion is met. We assume the pool-based scenario \cite{settles:2010}, in which the active learner has access to all unlabeled data. Since no labels are provided, and the percentage of sentences having at least one label is quite small, randomly sampling data is not an option, and AL is the obvious choice. Because it is easier for the human annotator to focus on only a single set of labels during the AL process, the text classification is realized using one independent classifier each for restrictions and topics (see Table \ref{tab:labels}). As the single labels under both restrictions and topics are not mutually exclusive, we train a multi-hot-encoded multi-label classification for both label sets.

%% file: acl21-sustainability-part5.tex
\section{Experiments} \label{sec:experiment}

We evaluate multi-label active learning performed by three human annotators, who each train a sentence classification model for classifying restrictions and topics, resulting in two runs per person. 

\subsection{Pre-processing and Experimental Setup}

Starting from the initial model, which is trained on the train set (described in Section~\ref{sec:data}), active learning is performed iteratively: \Ni 10 unlabeled sentences are presented to the annotator; \Nii The annotator may assign zero, one, or multiple labels per sentence; \Niii The newly-assigned labels are added to the train set, and a new model is trained. This process is repeated for 50 iterations.

\paragraph{Data} We use train, validation and test splits, as defined in Section \ref{sec:data_processing}, and an unlabeled pool consisting of 312,299 sentences. 

\paragraph{Query Strategy} For the query strategy, which selects the sentences to be labeled, we use prediction-entropy-based \cite{roy:2001} uncertainty sampling \cite{lewis:1994}, which selects the most uncertain samples, e.g., in this case those whose predicted class posterior exhibits the highest entropy. Since inference on transformers is computationally expensive, and we aim to keep the waiting times at a minimum, at the beginning of each iteration, we subsample the whole unlabeled pool randomly by selecting $4096$ examples \cite{mukherjee:2020}. Moreover, because the ratio of unlabeled sentences to sentences having at least one label is quite large, we adapt the query strategy to balance classes, by considering the class predictions and sampling evenly over the labels. In case this is not possible, e.g., when there is no prediction for a certain label, we fill the remainder with the remaining most uncertain samples, regardless of the predicted class.

\subsection{Model and Training}

Regarding the classification, we fine-tune the pre-trained \texttt{gbert-base} model \cite{chan:2020}, which has 110M parameters and is the best performing German transformer model for text classification at this number of parameters. While there is a larger gbert model available, we opted for the base variant due to its efficiency, which results in lower turnaround times of an AL step for the practitioner. We encode the labels as multi-hot encoded vectors. The model is trained using a softmax binary cross-entropy loss.

During each active learning iteration, the previous model is fine-tuned for $40$ epochs using a learning rate of \num{5e-5} on the data that has been labeled to this point. To avoid overfitting, we stop early when the validation loss has not changed for more than 5 epochs.

\subsection{Results}

Table \ref{tab:results1} shows the classification scores of aforementioned setting evaluated by three annotators and compared to an automated text classification baseline.

\input{table-results}

\begin{figure*}[b]
    \centering
    \includegraphics[width=\textwidth]{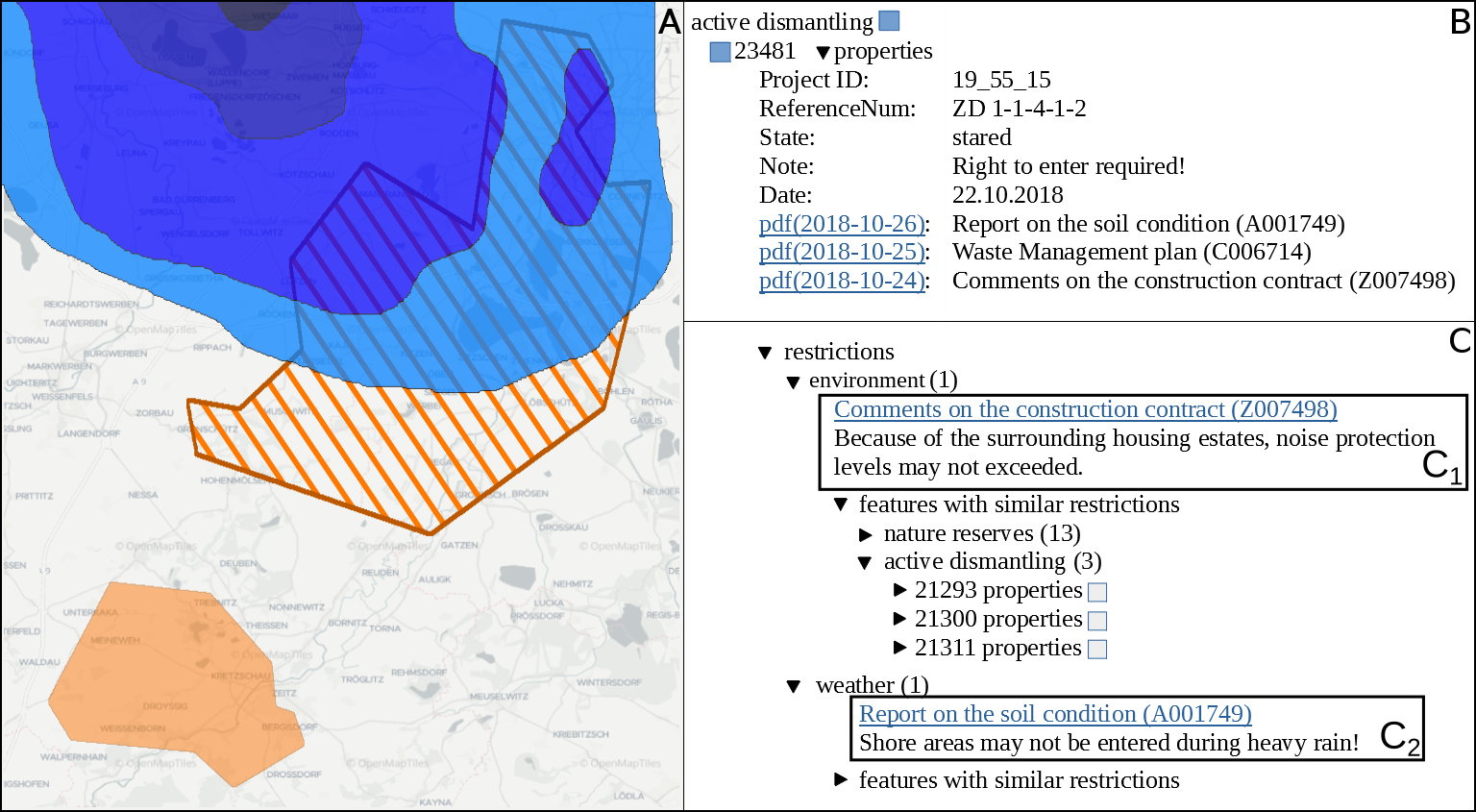}
    \caption{
    (A) Two geographic features of type "active dismantling" are displayed on a map. One feature was selected by mouse click (orange striped texture). The weather map is shown as isobands, with  precipitation values represented by shades of blue (dark blue tones indicate areas with high precipitation values). The information panel is displayed on the right hand side. It contains the non-spatial data of the selected geographic feature (B), as well as the usage restrictions together with a list of other features with similar usage restrictions (C).
    }
    \label{fig:gis}
\end{figure*}
The baseline is a \texttt{gbert-base} model trained on the initial data, i.e., without using AL at all. AL improves overall both micro-F1 and macro-F1 for topics by up to 3 percentage points, whereas improvements for restrictions seem marginal.

While the overall result improves just slightly, looking at the single labels, we can see considerable changes between plain text classification and active learning. Previously underperforming labels like ``Weather'' and ``Construction'' improve on average by 5 to 21 percentage points in F1. Smaller improvements can also be seen for ``Restricted Area'' and ``Environment'', and ``Disposal'' stays about the same. Unfortunately, ``Geotechnics'' and ``Planting'' and also drop in performance by 6 and 10 percentage points respectively. Interestingly, when we compare the difference in the relative quantities of co-occurring labels before and after the AL process, we find that the labeled pool changed notably during AL. We observed that \Ni the average number of labels per sentence increases; \Nii label co-occurrences shift considerably and some combinations even appear for the first time; \Niii every combination of topic labels occurs together in the data, which is not the case for our keyword-bootstrapped train set. (The exact numbers be seen in the Appendix, Figure \ref{app:fig:label_cooccurrence_initial_r}-\ref{app:fig:label_cooccurrence_diff}).

All in all, this indicates that AL is beneficial and improves classification metrics by a small amount, and moreover,  many samples with previously rarely or even unseen label combinations are found. Apparently, as these notable changes only lead to a small difference in F1, this new value of having more diverse label combinations is difficult to measure here against our keyword-bootstrapped test set. The only solution to a more representative test set, however, would require massive annotation efforts, since labels may be very sparse.

%% file: table-results.tex
\begin{table}[h!]
    \centering
    \small
    \begin{tabular}{l rp{0.01cm}lll r}
    \toprule
     & \textsc{F1 B.} & & \multicolumn{4}{c}{\textsc{F1 AL}} \\
     \cline{2-2} \cline{4-7}\rule{0pt}{2pt}\\
     \\[-1.9em]  
     \textsc{LABEL} & & &  A1 & A2 & A3 & \textsc{Avg.}\\
    \midrule
    \midrule
    \multicolumn{7}{c}{\textsc{\textbf{Restrictions}}}\\
    \midrule
    Prohibition & 0.93 & & 0.96 & 0.96 &  0.94& 0.95\\
    Requirement & 0.84 & & \underline{\bf 0.86} &  \underline{0.86} & 0.84  & 0.85\\
    \midrule
    \textsc{micro} & 0.87 & & 0.88 & 0.88 & 0.86 & 0.87\\
    \textsc{macro} & 0.90 & & 0.91 & 0.91 & 0.89 & 0.90\\
    \midrule
    \midrule
    \multicolumn{7}{c}{\textsc{\textbf{Topics}}}\\
    \midrule
     Weather & 0.53 & & 0.71 & 0.73 & \underline{0.77} & 0.74\\
    Construction & 0.58 & & \underline{\bf 0.63} & 0.64 & \underline{\bf 0.63} & 0.63\\
    Geotechnics & 0.58 & & \underline{\bf 0.50} & \underline{\bf 0.54} & \underline{\bf 0.53} & 0.52\\
    Restr. Area & 0.89 & & \underline{0.92} & \underline{\bf 0.91} & \underline{\bf 0.90} & 0.91\\
    Planting & 0.78 & & 0.73  & \underline{\bf 0.69} & \underline{\bf 0.61} & 0.68\\
    Environment & 0.73 & & 0.79  & 0.77 & 0.73 & 0.76\\
    Disposal & 0.73 & & \underline{\bf 0.72}  & 0.74 & 0.72 & 0.73\\
    \midrule
    \textsc{micro} & 0.70 & & 0.72 & 0.72 & 0.70 & 0.71\\
    \textsc{macro} & 0.69 & & 0.71 & 0.72 & 0.70 & 0.71\\
    \bottomrule
    \end{tabular}
    \caption{Active learning experiments, performed by three human annotators. ``AVG.'' is the annotator average over all three runs. ``F1 AL'' shows the final scores, broken down by annotator. ``F1 B.'' is a text classification baseline that is trained on the initial training set. For each label and annotator, we used McNemar's test \cite{mcnemar:1947} with $\alpha=0.05$ to test for significant change in the predictions compared to the baseline: We report obtained $p$-values, indicated by an underlined result for $p<0.05$, and bold text for $p<0.01$.}
    \label{tab:results1}
\end{table}

%% file: acl21-sustainability-part6.tex
\section{Visualization and Interaction Use Case}\label{sec:vis_usecase}

As an example, we present a workflow regarding areas which may not be entered during heavy rains for safety reasons. To answer a request (see Section~\ref{sec:data}), which e.g., is asking if a specific area may be entered, the expert uses the GIS, centers the map on the corresponding area, and displays the associated features (e.g., active dismantling areas, see Figure \ref{fig:gis}~A). To enable the expert to analyze the different feature categories, the displayed features are colored by category as suggested by \citet{Ware:2012}. Since areas can overlap in the map display, all features are colored only semi-transparently. 

Information immediately prohibiting certain activities is identified by clicking on a feature, which displays the non-spatial data in an information panel (Figure \ref{fig:gis}~A-B). To keep the expert's overview of the selected features, they are represented with a striped texture. All restrictions that result from the documents linked to the selected feature are listed (Figure \ref{fig:gis}~C). The entries are grouped by the restriction type and sorted by a confidence value (Figure \ref{fig:gis}~C\textsubscript{1} and C\textsubscript{2}). The document title and the sentence from which the restriction is derived are indicated. A click on the title opens a new window for reading the document. This list provides the expert with direct feedback on which documents might be relevant and, without reading them completely, an overview of which usage restrictions are present. This information is crucial for the experts, as it can have a significant impact on planned projects and their planning time. Additionally, the area described by a document can be superimposed with weather data. In this way, decisions regarding conditional restrictions (e.g., ``{\em may not enter shore areas during heavy rain}'') can also be made more quickly and directly on the basis of the system. The selected features overlap with a heavy rain area represented by isobands, therefore the request is directly answered and access to the area is currently prohibited (Figure \ref{fig:gis}~A).

However, for other restrictions, more information may be necessary, because experts often compare the region of interest with similar regions. Therefore, we provide a filter to highlight all features within the same restriction topic (Figure \ref{fig:gis}~C). By analyzing similar regions, the expert can derive recommendations for action, which might be necessary for the land reuse of an area. Recommendations for possible usage restrictions can also be derived in this way. Furthermore, this comparison can prevent actions from not being taken or from being taken too late, because the current information does not make them appear necessary, but it is clear from similar projects that they may nevertheless become necessary. This leads to a safe and quick reuse of regions maintained in that manner since precautions can be taken in advance. 

%% file: acl21-sustainability-sum.tex
\section{Conclusions and Future Work}

In this work, we have presented and evaluated a system which automates information requests related to the post-management of former open pit mines by leveraging unstructured and geospatial data. We used active learning for multi-label text classification to extract restrictions and topics from unstructured text in legacy documents and visualized the results using a GIS. As a result, targeted queries about restrictions and topics at specific geographic locations can be obtained much more efficiently, thereby speeding up the process of land reuse, which directly contributes to several SDGs. Further research is needed to shift recall towards 100\% to minimize false negatives, then correcting false positives in the system.

%% file: acl21-sustainability-ethics.tex
\section*{Ethical Considerations}

This work presents a workflow for the automatic information extraction in reports related to mining, construction and nature conservation. The collected information represents issues such as access restrictions or hazards. We are aware that misclassification in the application can lead to people being endangered or prevented from entering these regions for no reason. Misuse cannot be ruled out, but currently no specific example is known.

To ensure that misclassifications do not impact the stakeholders of the application, a quality assurance process will be used in the operating company so that employees in the piloting phase manually check where errors or information losses can be detected. In addition, there will be quality assurance for the application so that the probability of missing restrictions is minimized. Furthermore, our results could in theory lead to a decline in employees needed to read and check old documents, possibly resulting in job losses. Our scenario, however, requires specialists, who are not easily replaceable.

%% file: acl21-sustainability-supplementary.tex
\input{table-keywords}
\section{Data} 

\subsection{Keywords}

For each label, the keywords used to create the dataset are shown in Table~\ref{tab:keywords}.

\subsection{Inter-Annotator Agreement}

\input{table-iaa}

In Table~\ref{tab:iaa} we report Krippendorff's $\alpha$ and {Fleiss'~$\kappa$} for three human annotators.

\subsection{Absolute Label Occurrences}

\begin{figure}[ht]
\includegraphics[width=\linewidth]{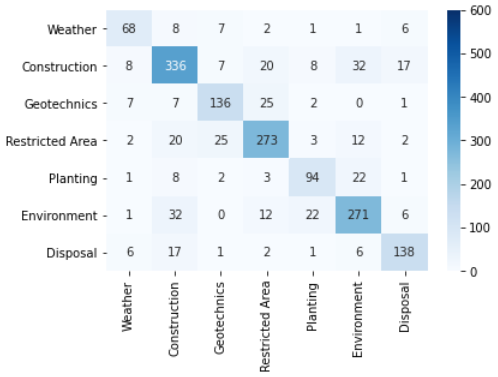}
\caption{Label co-occurrences.}
\label{app:fig:absolute_cooccurrences}
\vspace*{0.3cm}
\end{figure}

Table~\ref{tab:label_distribution} shows the absolute label distribution and Figure~\ref{app:fig:absolute_cooccurrences} shows the co-occurrence among labels.

\input{table-label-distribution}

~

\subsection{Relative Label Co-occurrence}

\begin{figure}[!ht]
\includegraphics[width=\linewidth]{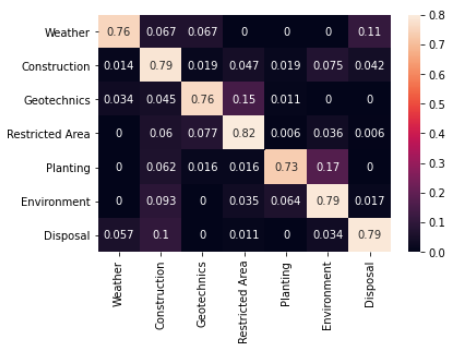}
\vspace*{0.3cm}
\caption{Samples found by keyword matching (labeled data).}
\label{app:fig:label_cooccurrence_initial_r}
\end{figure}

\begin{figure}[!ht]
\includegraphics[width=\linewidth]{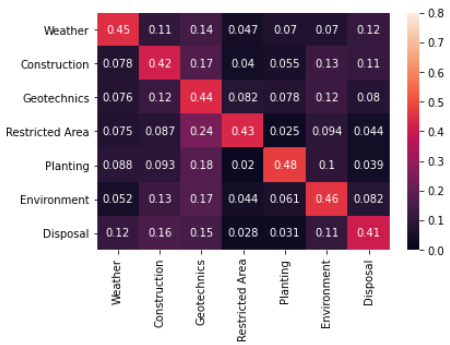}
\caption{Samples found by the query strategy.}
\label{app:fig:label_cooccurrence_query_strategy_r}
\end{figure}

\begin{figure}[!ht]
\includegraphics[width=\linewidth]{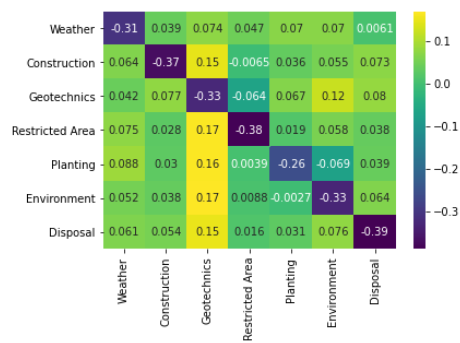}
\caption{Difference between the labeled data (after active learning) and the initial set.}
\label{app:fig:label_cooccurrence_diff}
\end{figure}

We show the relative label co-occurrence for the initial labeled set in Table~\ref{app:fig:label_cooccurrence_initial_r} (normalized per row). On the other hand, Table~\ref{app:fig:label_cooccurrence_query_strategy_r} shows the relative labels co-occurrence of the samples selected by the query strategy. The difference between those two Figures is shown by Figure~\ref{app:fig:label_cooccurrence_diff}.\\
\vspace*{0.3cm}

\newpage

%% file: table-keywords.tex
{
\renewcommand{\arraystretch}{1.4}
\begin{table*}[b]
    \centering
    \small
    \begin{tabular}{p{0.2cm} l p{6cm} p{6cm}}
        \toprule
        & \textsc{\bfseries{}Label} & \textsc{\bfseries{}Keywords} & \textsc{\bfseries{}English Translation}\\
        \hline
        \tablespacing
        \multirow{2}{*}[-0.15em]{\rotatebox{90}{\small\textbf{Restrictions}}} & \textbf{Prohibition} & 'verboten', 'nicht gestattet', 'nicht erlaubt', 'untersagt', 'unbefugt', 'darf nicht' & 'not permitted', 'not allowed', 'banned', 'unauthorized', 'may not' \\
        & \textbf{Requirement} & 'm\"ussen', 'muss', 'darf', 'nur', 'maximal', 'beachten' & 'must', 'must'(inflected), 'may', 'only', 'at most', 'consider'\\
        \tablespacing
        \hline
        \multirow{7}{*}[-2.5em]{\rotatebox{90}{\small\textbf{Topics}}} & \textbf{Weather} & 'Nebel', 'Wetter', 'Sturm', 'Starkniederschlag', 'Frost', 'Trockenheit', 'Regen', 'Schnee', 'Temperatur' & 'fog', 'weather', 'storm', 'heavy rainfall', 'frost', 'drought', 'rain', 'snow', 'temperature'\\
        & \textbf{Construction} & 'Bebauung', '\"uberbauung', 'errichten', 'Fenster', 'Mauer'&  'Construction', 'build on', 'construct', 'window', 'wall'\\        
        & \textbf{Geotechnics} & 'geotechnsch', 'Gel\"ande', 'Risse', 'Absenkung', 'Boden', 'Sohle' & 'geotechnical', 'terrain', 'crack', 'sinking', 'soil', 'horizon'\\      
        & \textbf{Restricted area} & 'Aufenthalt', 'Uferseitig', 'betreten', 'befahren', 'anlegen' & 'stay', 'shore-sided', 'enter', 'drive on', 'dock'\\  
        & \textbf{Planting} &'B\"aume', 'Baum', 'Pflanzen', 'f\"allen', 'forst' & 'trees', 'tree', ' plants', 'chop', 'forest'\\   
        & \textbf{Environment} &'Nester', 'Arten', 'Umwelt', 'gesch\"utzt' &'nests', 'species', 'environment', 'protected'\\   
        & \textbf{Disposal} & 'lager', 'entsorg', 'abfall', 'verbringen', 'verklappen' & 'store', 'disposal', 'waste', 'remove', 'dumping'\\   
        \bottomrule
    \end{tabular}
    \caption{Keywords used for dataset generation (in German) and their English translation.}
    \label{tab:keywords}
\end{table*}
}

%% file: table-iaa.tex
\begin{table}[!ht]
\begin{tabular*}{\linewidth}{@{}lrr@{}}
\toprule
\makecell[lb]{\textsc{\textbf{Label}}\\~} & \makecell[lb]{\textsc{\textbf{Krippen-}}\\\textsc{\textbf{dorff's $\alpha$}}} & \makecell[lb]{\textsc{\textbf{Fleiss' $\kappa$}}\\~} \\ 
\midrule
Prohibition     & 0.8988         & 0.8995              \\ 
Requirement     & 0.8303        &     0.8317          \\ 
Weather         & 0.7400        &     0.7506          \\ 
Construction    & 0.6991        &     0.7010        \\ 
Geotechnics     & 0.7095        &     0.7150          \\ 
Restricted area & 0.9140        &     0.9143         \\ 
Planting        & 0.7579        &     0.7653          \\ 
Environment     & 0.7542        &     0.7555          \\ 
Disposal        & 0.8118        &     0.8138          \\ 
\bottomrule
\end{tabular*}
   \caption{Krippendorff's alpha and Fleiss' kappa for each label, each sample in the dataset was annotated by three different annotators.}
    \label{tab:iaa}
\end{table}

%% file: table-label-distribution.tex
\begin{table}[!ht]
\begin{tabular*}{\linewidth}{l@{}rrrr}
\toprule
\textsc{\textbf{Label}}  & \textsc{\textbf{train}} & \textsc{\textbf{test}} & \textsc{\textbf{val}} & \textsc{\textbf{total}}\\ \midrule
Prohibition     & 47             & 93            & 47  &  187        \\
Requirement     & 149            & 299           & 149  &  597       \\
Weather         & 17             & 34            & 17   &  68       \\
Construction    & 84             & 168           & 84    & 336       \\
Geotechnics     & 34             & 68            & 34 &  136        \\
Restricted area & 69             & 136           & 68   & 273        \\
Planting        & 23             & 47            & 24  & 94         \\
Environment     & 68             & 135           & 68  &   271       \\
Disposal        & 34             & 69            & 35  &  138       \\ \bottomrule
\end{tabular*}
   \caption{Label distribution in the train-, test-, and validation data set}
    \label{tab:label_distribution}
\end{table}